\documentclass[conference, 9pt]{IEEEtran}
\IEEEoverridecommandlockouts
\usepackage{cite}
\usepackage{amsmath,amssymb,amsfonts}
\usepackage{algorithmic}
\usepackage{graphicx}
\usepackage{textcomp}
\usepackage{xcolor}
\usepackage{subfig}
\usepackage[numbers,sort&compress]{natbib}
\def\BibTeX{{\rm B\kern-.05em{\sc i\kern-.025em b}\kern-.08em
    T\kern-.1667em\lower.7ex\hbox{E}\kern-.125emX}}
\begin{document}
\title{Multiplex Graph Contrastive Learning with Soft Negatives
\thanks{\textsuperscript{\dag} These two authors contributed equally.}
\thanks{\textsuperscript{*} Correspondence should be addressed to: wrcai@suda.edu.cn or chenl\\@snnu.edu.cn}
}


\author{Zhenhao Zhao\textsuperscript{1,\dag}\qquad Minhong Zhu\textsuperscript{2,\dag}\qquad Chen Wang\textsuperscript{1}\qquad Sijia Wang\textsuperscript{1}\qquad Jiqiang Zhang\textsuperscript{4}\qquad Li Chen\textsuperscript{3,*}\qquad Weiran Cai\textsuperscript{1,*} \\  \IEEEmembership{\textsuperscript{1}School Of Computer Science and Technology, Soochow University, Suzhou, China}\\
\IEEEmembership{\textsuperscript{2}School of Biology and Basic Medical Sciences, Soochow University, Suzhou, China}\\
\IEEEmembership{\textsuperscript{3}School of Physics and Information Technology, Shaanxi Normal University, Xi‘an, China}\\
\IEEEmembership{\textsuperscript{4}School of Physics, Ningxia University, Yinchuan, China}
}

\maketitle
\begin{abstract}
Graph Contrastive Learning (GCL) seeks to learn nodal or graph representations that contain maximal consistent information from graph-structured data. While node-level contrasting modes are dominating, some efforts commence to explore consistency across different scales. Yet, they tend to lose consistent information and be contaminated by disturbing features. Here, we introduce MUX-GCL, a novel cross-scale contrastive learning paradigm that utilizes multiplex representations as effective patches. While this learning mode minimizes contaminating noises, a commensurate contrasting strategy using positional affinities further avoids information loss by correcting false negative pairs across scales. Extensive downstream experiments demonstrate that MUX-GCL yields multiple state-of-the-art results on public datasets. Our theoretical analysis further guarantees the new objective function as a stricter lower bound of mutual information of raw input features and output embeddings, which rationalizes this paradigm. Code is available at https://github.com/MUX-GCL/Code.
\end{abstract}


\vspace{-0.5em}
\section{Introduction}
\vspace{-0.2em}
Taming graph-structured data has been one of the major challenges in machine learning, which is coined as graph representation learning (GRL). While significant progresses have been made, notably with paradigms incorporating both nodal and topological information, most prevailing methods are supervised learning \cite{kipf2016semi-GCN,velivckovic2017GAT,wu2019simpGCN,xu2018powerfulGNN,xu2019GNN}. Yet, GRL has not only to face a majority of graph data for which labels are unavailable in real-world scenarios, but more desirably to discover patterns in an autonomous way \cite{chen2020simple}. To tackle this challenge, recent studies have extensively explored the realm of self-supervised learning (SSL), among which graph contrastive learning (GCL) plays a pivotal role.

In essence, GCL aims to learn nodal or graph representations by maximizing the information consistency between augmented views of the graph. Most of the established methods share the spirit of operating same-scale contrast between nodal representations through on positive and negative pairs \cite{zhu2020GRACE, you2020graphCL, zhu2021GCA}. For graph-structured data, however, feature consistency can be well conveyed in structures of different scales \cite{velivckovic2018DGI}. Some efforts have thus expanded the scope to cross-scale modes, including \textit{patch-global} contrast of nodal and graph representations \cite{velivckovic2018DGI, hassani2020MVGRL, mavromatis2021GIC}, and \textit{context-global} contrast between contextual subgraph- and graph-levels \cite{cao2021BIGI, wang2021HTC}. The contrasts of patches at diverse scales prove to be highly beneficial.  

Yet, with the gain of richer information, cross-scale contrasting modes tend to suffer from contamination by inconsistent features.
The expansion to larger-scale patches tends to join out-of-class nodes and hence more feature inconsistency. It is thus an intriguing question: \textit{How to enable contrasts that capture more consistent features across scales while restrict contamination from inconsistency?}

This raises a request for a contrasting paradigm that exploits information maximally and selectively. One has to note that information loss is inherent in GCL. On one hand, an encoding process is not guaranteed information-conservative. The inclination for oversmoothing is intrinsic to message-passing based methods. On the other hand, pairing negatives between intra-class nodes leads to a loss of consistent features. This has been spotted in the same-scale contrast. Regarding this, some work excludes neighbouring nodes to avoid false negatives \cite{xia2021progcl, niu2023AUGCL} or weighs them as positives based on their saliency \cite{li2023homogcl}. However, these approaches are not applicable to topological compositions in cross-scale scenarios.

\begin{figure}
\vspace{-1.6em}
\centering
\includegraphics[width=2.6in]{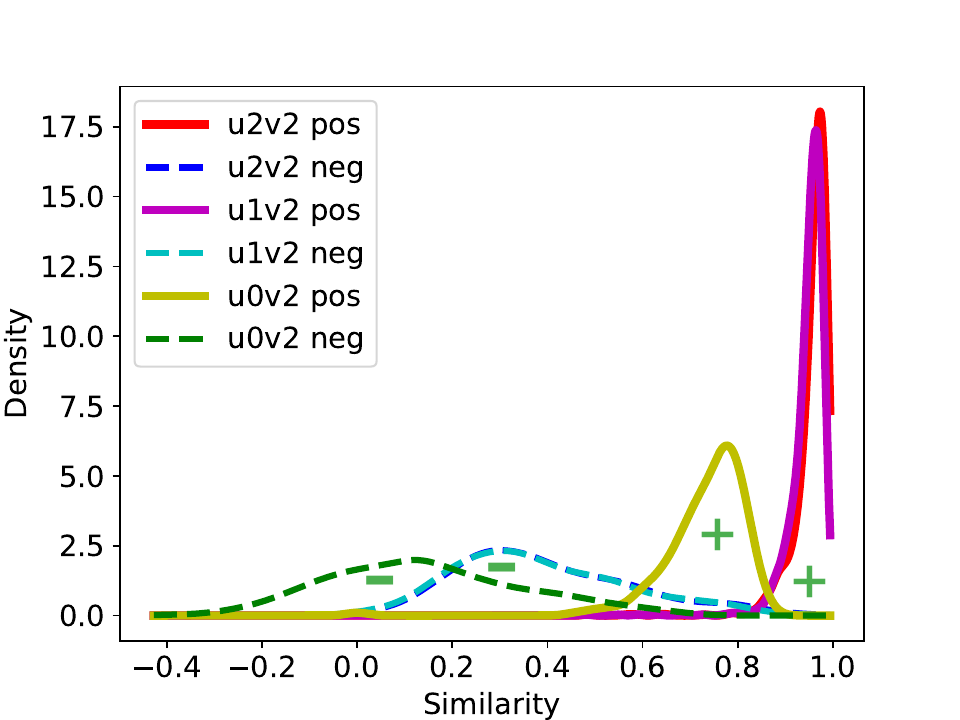}
\vspace{-0.2em}
\caption{Similarity distributions of cross-layer embeddings between two augmented views (for GRACE).  All positive pairs are substantially more similar than negative pairs, labeled as $u_mv_n~pos/neg$ with $m$ and $n$ being the layers. }

\label{cls}
\vspace{-1.8em}
\end{figure}

We propose MUX-GCL, a novel cross-scale contrastive learning paradigm that puts multiplex encoded information into full play. The core of the paradigm lies in the contrasts of ``effective patches" constructed from all latent representations of the encoder. Concretely, higher-layer nodal embeddings, interpreted as representations of patches centered on focal nodes, are contrasted with lower-layer embeddings, where features are less contaminated by the locality. To assist such cross-scale contrasts, a multiplex contrasting strategy is proposed to minimize information loss from false negative pairs, guided by topological affinities of patches. With these facilities, consistent information contained in the entire multiplex encoder is maximally exploited.  
\begin{figure*}
\vspace{-1em}
    \centering
      {\includegraphics[width=5.5in]{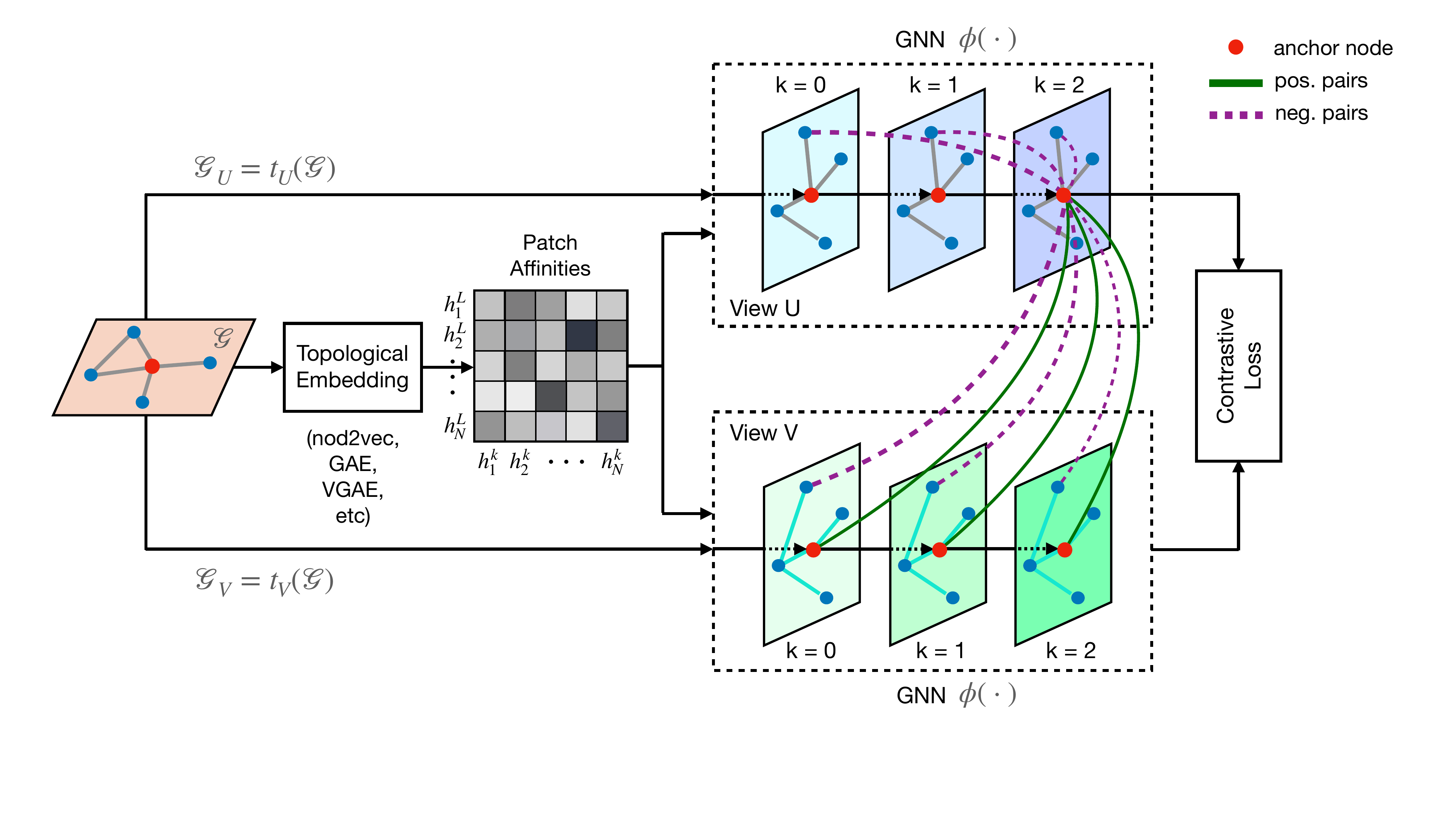}}
      \vspace{-3.5em}
    \caption{Overall architecture of MUX-GCL. Contrasts are executed between ``effective patches" constructed from all representations of the multiplex encoder, as illustrated by the links. The pairwise affinities of topological embedding estimate the likelihood of being false negatives. Augmentations are implemented as in GRACE.}
    \label{overview}
    \vspace{-2em}
\end{figure*} 

Our contributions are summarized as follows:
\begin{itemize}
\vspace{-0.2em}
    \item We propose a novel cross-scale GCL paradigm, MUX-GCL, utilizing multiplex representations of the entire encoder to maximize consistent information while mitigate disturbing features.
    \item We introduce a patch contrasting strategy based on topological affinities to alleviate false negative pairs, which most notably, is the only mechanism applicable to cross-scale contrasts.
    \item Our theoretical justification guarantees the objective function of MUX-GCL as a stricter lower bound of mutual information between raw features and learned representations of augmented views, providing the rationale behind the method.
    \item Extensive experiments on both classification and clustering tasks demonstrate salient improvements, outperforming multiple state-of-the-art GCL models on public datasets.  
\end{itemize}

\vspace{-0.2em}
\section{Methods}
\vspace{-0.2em}
\subsection{Preliminaries and Notations}\label{Preliminaries}
\vspace{-0em}
Let $\mathcal{G} =  ( \mathcal{V,\mathcal{E} }  ) $ denote a graph, where $\mathcal{V}=\left \{ v_i \right \}_{i=1}^N $, $\mathcal{E} \subseteq \mathcal{V} \times \mathcal{V}$ are the node and edge sets respectively. We let $ X\in \mathcal{R} ^{ N\times F } $ and $A\in \left \{ 0, 1 \right \}^{ N\times N} $ be the feature matrix and adjacency matrix. 
As a form of SSL, the purpose of our model is to learn a reliable representation $f ( X,A  ) \in \mathcal{R} ^{N\times F}$ of the input data with no labels through a GCN encoder. It is essential to support downstream tasks, such as node classification and clustering. Hence, the learned representations will be commonly input to a minimal prediction head for tests.


\vspace{-0em}
\subsection{Motivation}
\vspace{-0.0em}

We seek to establish a cross-scale contrastive learning method that gains richer consistent information. Two aspects are of our concern: the construction of multi-scale patches and the contrasting strategy. 

The key issue with conventional ways of constructing patch representations through pooling is the information loss caused by involving inconsistent features. Instead, we consider using the entire ensemble of latent and final representations of an encoder for building patches. From the perspective of message passing, we regard a $k$th-layer embedding of an anchor node as a representation of a $k$-hop ego-net centered on it, which forms an ``effective patch". This treats the encoder as a multiplex network, which introduces no extra information contamination.

Cross-scale contrasts may thus be established between pairs of such patch representations. The roles of such effective patches in contrasting can be justified by observing the similarity between cross-layer embeddings, as demonstrated for GRACE. As shown in Fig. \ref{cls}, all positive patch pairs, regardless of layers (scales), are far more similar than negative pairs, as suggested by the well separated distribution of similarities. This strongly indicates that representations across layers deserve to be involved in graph contrastive learning. 

Yet, to systematically pairing cross-scale patches, we need a contrasting strategy that maximally preserve consistent features. This aims essentially to avoid brutal erasure of exploitable information by pairing false negatives. In the absence of class labels, we evaluate the likelihood of false negatives on the topological affinities of patches as priors. The use of affinities thus builds up ``soft negatives" in contrast to the commonly used hard ones in GCL. 

\vspace{-0.5em}
\subsection{Framework}
\vspace{-0.2em}
From the rationale above, we establish ``effective patches" using all representations of the encoder for contrastive learning. Each nodal embedding $U^{(k)}$ ($V^{(k)}$ in the other view) on the $k$th layer of the GCN now serves as an effective representation of a $k$-hop ego-net centered at the anchor node. Specifically, the definition of the patch representation takes the standard form $U^{(k)}=\sigma (\tilde{A}U^{(k-1)}W^{(k)})\in\mathcal{R}^{N\times F}$ with the initial input $U^{(0)}=X$,
where $\tilde{A}$ is the normalized adjacency matrix and $W^{(k)}$ a set of trainable parameters. 


With this premise, we now introduce the cross-scale contrastive learning paradigm MUX-GCL (Fig. \ref{overview}). We deploy two modules, i.e. \textbf{multiplex patch contrast} and \textbf{patch affinity estimation}.\\
\vspace{-1em}

\noindent\textbf{Multiplex Patch Contrast (MPC)}.
To contrast effective patches across scales, we extend the commonly used InfoNCE loss from same-scale contrast to a multiplex setting. Since final representations of the encoder are ultimately desired, we conduct cross-scale contrasts between final and all intermediate layers representations. The multiplex objective function is given as follows
\vspace{-0.5em}
\begin{equation}\label{InfoNCE}
\begin{array}{lc}
\mathcal{L}_{c} 
(u_i^{(L)},v_i^{(k)}) = \\
\log \frac{{{{\rm{e}}^{\theta ( {{u_i^{(L)}},{v_i^{(k)}}} )/\tau }}}}
{{{{\rm{e}}^{\theta ( {{u_i^{(L)}},{v_i^{(k)}}} )/\tau }} 
+\sum\limits_{j \ne i}\omega_{ij}^{Lk}  {{{\rm{e}}^{\theta ( {{u_i^{(L)}},{v_j^{(k)}}}
)/\tau }}}  + \sum\limits_{j \ne i} \omega_{ij}^{Lk} 
 {{{\rm{e}}^{\theta ( {{u_i^{(L)}}, {u_j^{(k)}}} )/\tau }}} }}  
\end{array}
\end{equation}

\noindent where $\theta (\cdot,\cdot)$ is the similarity function. The metric $\omega_{ij}^{Lk}$ represents a measure of the likelihood of being false negatives. To treat the contrasts in a balanced way, we average the objective function across different scales, as expressed by the pairwise objective function
\begin{equation}\label{lambda}
\mathcal{L}_{c}(u_i,v_i) = \sum_{k=0}^{L}\lambda_k \mathcal{L}_{c} (u_i^{(L)},v_i^{(k)})
\end{equation}
where $\lambda_k$ is the weight for contrasting the final $L$-th layer and the intermediate $k$-th layer, with $\sum_{k=0}^{L} \lambda_k=1$. 

Finally, to ensure symmetry between the two views, the overall objective function is defined as
\vspace{-0.2em}
\begin{equation}
    \mathcal{L}_{MUX} =\frac{1}{2N} \sum^{N}_{i=1}[\mathcal{L}_c (u_i,v_i)+ \mathcal{L}_c (v_i,u_i)]. 
\end{equation}

\noindent\textbf{Patch Affinity Estimation (PAE)}\label{MA}. The affinity estimation function assigns weights to negative pairs to alleviate the problem of false negatives. Notably, in the cross-scale contrast, patches are more likely to share information due to their positional affinity, where overlaps are significantly more incident. A higher affinity score thus indicates a higher likelihood of being false negatives. This weighting scheme is thus to reduce the loss of consistent information in negative pairs. 

For this scenario, we propose an affinity estimation strategy using topological positions as a decent prior. Concretely, we employ a graph embedding algorithm to obtain nodal representations that contain solely topological information. The topological representation of a patch is then simply obtained by pooling the encompassed nodes
\vspace{-0.2em}
\begin{equation}
    H^{(0)}=T(A, X) \qquad
    h_i^{(k)} = Pool_{j\in G_i^{(k)}}(h_j^{(0)})
\end{equation}
where $T(\cdot)$ represents a learning algorithm that maps nodes to a topological embedding space. $G_i^{(k)}$ represents the $k$-hop ego-network centered on node $i$. $Pool$ denotes the pooling function aggregating nodal embeddings within the patch.

Here we consider two learning algorithms to obtain the topological embeddings: Node2Vec \cite{grover2016node2vec} and VGAE (Variational Graph Auto-Encoder) \cite{kipf2016VGAE}.
We remark that the decoder of VGAE is to recover the adjacency matrix of the input graph and hence learns topological features only. 

To obtain the inter-patch affinities, we compute the similarities of these topological representations. Based on the affinity score for a negative instance pair, we compute the weight $\omega$ as the estimated likelihood of being false negatives
\begin{equation}\label{affinity}
\omega_{ij}^{Lk} =1-\eta(h_i^{(L)},h_j^{(k)})
\end{equation}
where $k\in \{0,1,\dots , L\}$; $\eta(\cdot,\cdot)$ is the affinity function that measures the positional similarity. Here, we take the form of normalized inner product $\eta(h_i^{(L)},h_j^{(k)})=\langle h_i^{(L)}, h_j^{(k)}\rangle$.

\vspace{-0em}
\subsection{Theoretical Justification}
\vspace{-0em}
We provide a theoretical justification for our proposed multiplex contrastive objective, demonstrating its rationale through the lens of maximization of mutual information.

\textbf{Proposition 1.} \textit{The multiplex contrastive objective in Eq.\ref{InfoNCE} is a lower bound of mutual information (MI) between raw input features $\mathbf{X}$ and output node embeddings $\mathbf{U}$ and $\mathbf{V}$ in the two augmented views. Further, with a statistical significance, the objective is also a stricter lower bound compared with the contrastive objective $\mathcal{L}_{GR}$ proposed by the benchmark $GRACE$. Formally,}
\begin{equation}
    \mathcal{L}_{GR} < \mathcal{L}_{MUX} < I(\mathbf{X};\mathbf{U,V}).
\end{equation}
\noindent Proof. We first prove $\mathcal{L}_{MUX} < I(\mathbf{X;U,V})$. Let $\mathbf{U}^{(k)}, \mathbf{V}^{(k)}$ (for $k=0, 1, \dots, L$) be the embeddings generated by the $k$-th layer of the encoder. Our proposed objective includes $2(L+1)$ cross-scale contrasting pairs
\begin{eqnarray}
\label{L_mux}
    {\mathcal{L}_{MUX}}  = \frac{1}{2}\sum\limits_{k = 0}^L \frac{\lambda_k}N \sum\limits_{i = 1}^N \left[ \mathcal{L}_c(u_i^{(L)},v_i^{(k)}\right) + \mathcal{L}_c(v_i^{(L)},u_i^{(k)} )]
\end{eqnarray}
For sufficiently large $N$, we have ${\omega_{ij}^{Lk}} > 1/N$, which renders 
\begin{eqnarray}
 I_{NCE}(\mathbf{U}^{(L)},\mathbf{V}^{(k)}) > \frac1N \sum\limits_{i = 1}^N \mathcal{L}_c(u_i^{(L)},v_i^{(k)}).
\end{eqnarray}
As InfoNCE is a lower bound of $MI$, we have 
\begin{eqnarray}
{\mathcal{L}_{MUX}} < \frac{1}{2}\sum\limits_{k = 0}^L\lambda_k
\left[I(\mathbf{U}^{(L)};\mathbf{V}^{(k)}) + I(\mathbf{V}^{(L)};\mathbf{U}^{(k)}\right)].
\end{eqnarray}
Resorting to the relations $I(\mathbf{U}^{(L)};\mathbf{V}^{(k)}) \le I(\mathbf{X};\mathbf{U}^{(L)}) = I(\mathbf{X};\mathbf{U}) \le I(\mathbf{X};\mathbf{U},\mathbf{V})$ \cite{zhu2020GRACE} and noticing the normalized $\lambda_k$, we finally have
\begin{eqnarray}
\mathcal{L}_{MUX} < \frac{1}{2}\sum\limits_{k = 0}^L\lambda_k\left[I(\mathbf{X};\mathbf{U},\mathbf{V}\right) + I(\mathbf{X};\mathbf{V},\mathbf{U})] = I(\mathbf{X};\mathbf{U},\mathbf{V}).
\end{eqnarray}

We then show that $\mathcal{L}_{MUX} > \mathcal{L}_{GR}$ with a statistical significance. We first rewrite the loss function of GRACE as \begin{eqnarray}
\label{L_GR}
\mathcal{L}_{GR} = \frac12\sum\limits_{i = 1}^N \left[\mathcal{L}_g(u_i^{(L)},v_i^{(k)}\right) + \mathcal{L}_g(v_i^{(L)},u_i^{(k)} )]
\end{eqnarray}
with $\mathcal{L}_g(u_i^{(L)},v_i^{(L)}) = \left[1+\sum\limits_{j \ne i}(\rm{e}^{\psi_{S,ij}^{Lk}} + \rm{e}^{\psi_{D,ij}^{Lk}})\right]^{-1}$,
where $\psi_{S,ij}^{Lk}=\theta (u_i^{(L)},u_j^{(k)})-\theta (u_i^{(L)},v_i^{(k)})$ and $\psi_{D,ij}^{Lk}=\theta (u_i^{(L)},v_j^{(k)})-\theta (u_i^{(L)},v_i^{(k)})$. Similarly, the loss function in $\mathcal{L}_{MUX}$ can be written as $\mathcal{L}_c(u_i^{(L)},v_i^{(k)})=\left[1+\sum\limits_{j \ne i}\omega_{ij}^{Lk}(\rm{e}^{\psi_{S,ij}^{Lk}} + \rm{e}^{\psi_{D,ij}^{Lk}})\right]^{-1}$.

\begin{figure}[tbp]
  \centering
  {\includegraphics[width=0.242\textwidth]{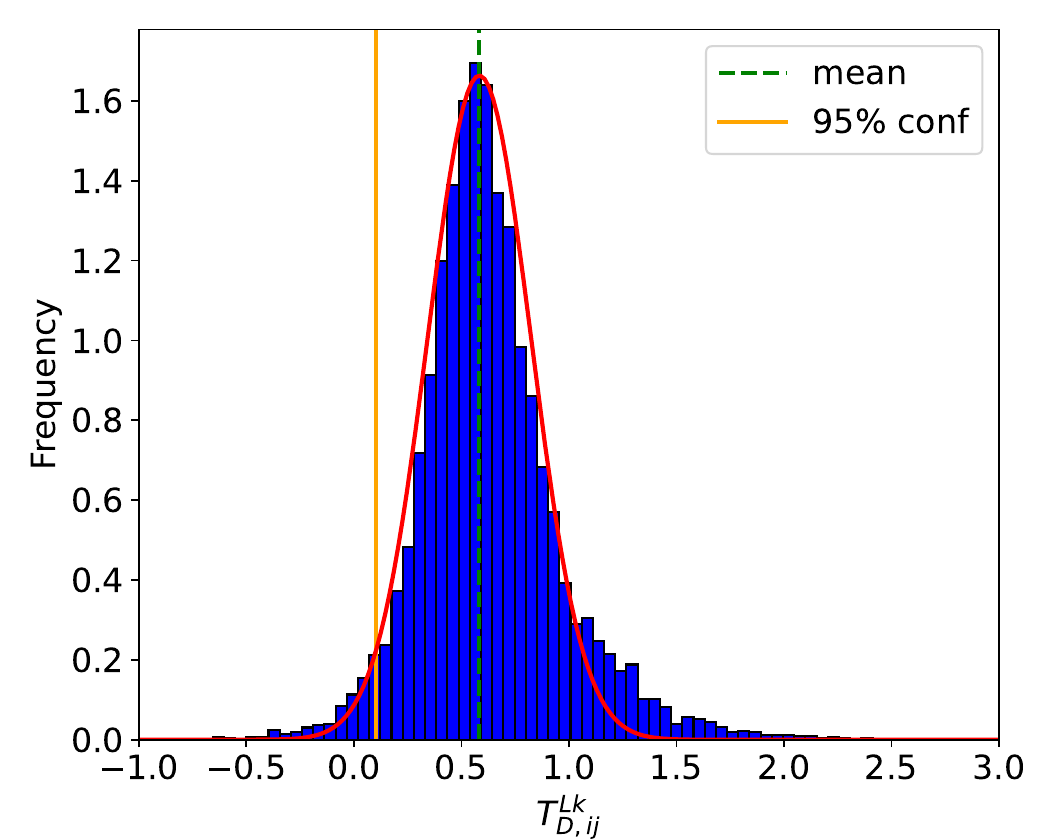}}
  {\includegraphics[width=0.24\textwidth]{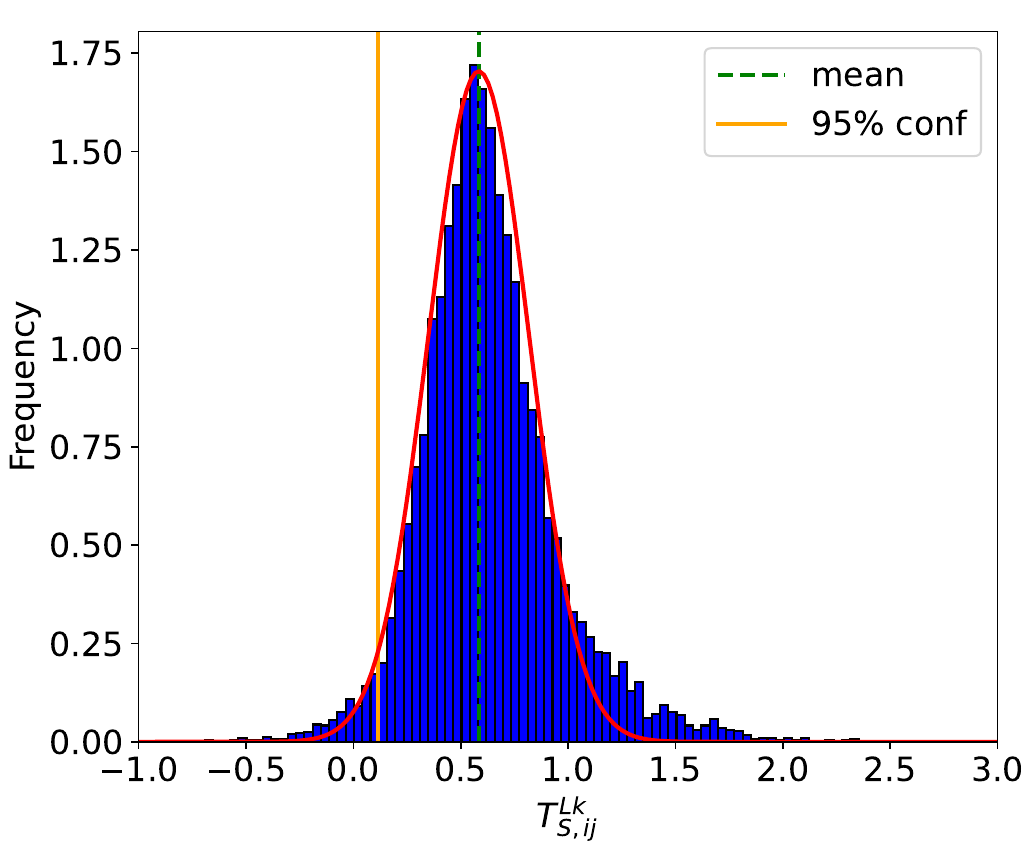}}
  \caption{Distributions of $T_{D,ij}^{Lk}$ (left) and $T_{S,ij}^{Lk}$ (right) for Cora fitted by Gaussian curves. Results are shown for epoch 300, but are consistent for the entire training process.}
  \label{curve}
  \vspace{-1em}
\end{figure}

To compare $\mathcal{L}_c$ and $\mathcal{L}_g$, we define
$T_{S,ij}^{Lk}=\psi_{S,ij}^{Lk}-\psi_{S,ij}^{LL}+\log \omega_{ij}^{Lk}$ and $
T_{D,ij}^{Lk}=\psi_{D,ij}^{Lk}-\psi_{D,ij}^{LL}+\log \omega_{ij}^{Lk}$. From the statistics as shown in Fig \ref{curve}, we show that throughout the training, both quantities, well fitted by Gaussian distribution, are positive with a great statistical significance (within the 95\% confidence interval). We can thus conclude that with a large probability, $\mathcal{L}_c(u_i^{(L)},v_i^{(k)})>\mathcal{L}_g(u_i^{(L)},v_i^{(L)})$; symmetrically, $\mathcal{L}_c(v_i^{(L)},u_i^{(k)})>\mathcal{L}_g(v_i^{(L)},u_i^{(L)})$. These relations also hold for $k=L$ since $\omega_{ij}^{LL} \in (0,1)$ for $j\ne i$. Hence, by comparing the entire expressions, we finally reach $\mathcal{L}_{MUX}>\mathcal{L}_{GR}$. 

We can hence conclude that maximizing $\mathcal{L}_{MUX}$ is equivalent to maximizing a lower bound of the mutual information between raw features and learned node representations, which is yet stricter than the commonly used contrastive objective. It guarantees the convergence and provides a theoretical base for the performance.
\vspace{-0.2em}
\subsection{Time Complexity Analysis}\label{time_analysis}
\vspace{-0.2em}
The time cost of the multiplex contrast mechanism is limited compared to the prevailing GCL methods. Concretely, we choose GRACE for comparison. Given a graph with $N$ nodes and $E$ edges, and assuming a GCN encoder with $L$ layers and $d$ hidden dimensions, the time complexity of encoding and loss function of GRACE are $O(L(Nd^2+Ed))$ and $O(N^2d)$, respectively. For the encoding stage, MUX-GCL takes extra $O(LNd^2)$ to acquire intermediate embeddings through linear layers, which does not increase the time complexity significantly, as $L$ is typically very small ($L=2$ for most cases). For the loss function, the time complexity of MUX-GCL is $O((L+1)N^2d)$, which is on the same order of magnitude as GRACE, noting that the InfoNCE loss in GRACE is a special case of Eq. \ref{lambda} when $\lambda_L=1$. Furthermore, the time complexity of Node2Vec and VGAE used in the PAE module are $O(N)$ and $O(Nd^2+Ed)$. This does not add to the overall complexity since PAE can be implemented as pre-processing and computed only once in the training phase.

\vspace{-0.5em}
\section{Experiments}
\vspace{-0.5em}
\subsection{Experimental Setup}
\vspace{-0.2em}
\begin{table}[]
\setlength{\tabcolsep}{4pt}
    \caption{ Node classification results (Acc (\%) ± Std for 5 seeds).}
    \vspace{-0.8em}
    \label{main_results}
    \centering
    \begin{tabular}{lcccccc}
    \hline
    \tabcolsep=1em
        Model  & Cora & Citeseer & Pubmed & Photo & Comp. \\ \hline
        raw features & 64.8±0.1 & 64.6±0.1 & 84.8±0.0 & 78.5±0.0 & 73.8±0.0 \\ 
        node2vec\cite{grover2016node2vec} & 74.8±0.0 & 52.3±0.1 & 80.3±0.1& 89.7±0.1 & 84.4±0.1 \\ 
        DeepWalk\cite{perozzi2014deepwalk} & 75.7±0.1 & 50.5±0.1 & 80.5±0.2 & 89.4±0.1 & 85.7±0.1 \\ 
        GAE\cite{kipf2016VGAE} & 76.9±0.0 & 60.6±0.2 & 82.9±0.1 & 91.6±0.1 & 85.3±0.2  \\ 
        VGAE\cite{kipf2016VGAE} & 78.9±0.1 & 61.2±0.0 & 83.0±0.1 & 92.2±0.1 & 86.4±0.2 \\ 
        DGI\cite{velivckovic2018DGI} & 82.6±0.4 & 68.8±0.7 & 86.0±0.1 & 91.6±0.2 & 84.0±0.5 \\ 
        GRACE\cite{zhu2020GRACE} & 83.3±0.4 & 72.1±0.5 & 86.3±0.1 & 92.5±0.2 & 87.8±0.2 \\ 
        MVGRL\cite{hassani2020MVGRL} & 83.8±0.3 & 73.1±0.5 & 86.3±0.2 & 91.7±0.1 & 87.5±0.1 \\ 
        GCA\cite{zhu2021GCA} & 82.8±0.3 & 71.5±0.3 & 86.0±0.2 & 92.2±0.2 & 87.5±0.5 \\ 
        SUGRL\cite{mo2022SuGRL} & 83.4±0.5 & 73.0±0.4 & 84.9±0.3 & 93.2±0.4 & 88.8±0.2 \\ 
        BGRL\cite{thakoor2021BGRL} & 83.7±0.5 & 73.0±0.1 & 84.6±0.3 & 91.5±0.4 & 87.3±0.4 \\
        G-BT\cite{bielak2022G-BT} & 83.6±0.4 & 72.9±0.1 & 84.5±0.1 & 92.6±0.5 & 86.8±0.3 \\
        ProGCL\cite{xia2021progcl} & 84.2±0.5 & 72.2±0.2 &86.4±0.2& 93.2±0.1 & 88.7±0.1 \\ 
        COSTA\cite{zhang2022costa} & 84.3±0.2 & 72.9±0.3 & 86.0±0.2 & 92.6±0.5 & 88.3±0.1 \\ 
        SFA\cite{zhang2023SFA} & 84.1±0.1 & 73.7±0.2 & 85.6±0.1 & 92.8±0.1 & 88.1±0.1 \\
        HomoGCL\cite{li2023homogcl} & 84.9±0.2 & 71.7±0.3 & 85.8±0.1 & 93.0±0.2 & 89.0±0.1 \\ 
        MA-GCL\cite{gong2023magcl} & 83.9±0.1 & 72.1±0.4 & 85.6±0.4 & 93.4±0.1
 & 89.0±0.1 \\ \hline
        MUX-GCL & \textbf{85.5±0.3} & \textbf{73.8±0.2} & \textbf{86.9±0.2} & \textbf{93.9±0.1} & \textbf{90.7±0.1} \\ \hline
    \end{tabular}
    \vspace{-2em}
\end{table}

\noindent\textbf{Datasets and Baselines}. We evaluate our method on five public available datasets ranging from citation networks and e-commercial sites: Cora, Citeseer, Pubmed, Amazon-Photo and Amazon-Computers. To be consistent with the previous GCL methods (GRACE, GCA, COSTA, SFA etc.), all datasets are randomly divided into 10\%, 10\%, and 80\% proportions for training, validation, and testing. We compare MUX-GCL with multiple baselines.\\
\noindent\textbf{Evaluation protocol}. Adhering to the evaluation framework used by prior work \cite{velivckovic2018DGI, zhu2020GRACE, zhu2021GCA}, we employ a standard two-layer GCN encoder and yield embeddings for downstream tasks. For the node classification task, we employ an $\ell_2$-regularized logistic regression classifier from the Scikit-Learn library \cite{pedregosa2011scikit}. For node clustering task, we employ KMeans as clustering method and measure the performance in terms of Normalized Mutual Information (NMI) score and Adjusted Rand Index (ARI) \cite{RI, li2023homogcl}.

\vspace{-0.5em}
\subsection{Node Classification}
\vspace{-0.2em}
 First, MUX-GCL surpasses all same-scale GCL methods, including advanced models like ProGCL, and HomoGCL, where only output embeddings are contrasted, whereas our paradigm forms patches with both latent and final representations. Second, it outperforms the methods that do not discern false negatives (e.g. GRACE, GCA, BGRL, COSTA) by assigning affinity-informed weights, which minimizes the loss of consistent information. Third, MUX-GCL is superior to previous cross-scale GCL methods (e.g. DGI, MVGRL) where larger-scale representations are typically pooled from final embeddings. Thanks to the use of all representations within the encoder for constructing patches, MUX-GCL contains less inconsistent information. The performance comparison is summarized in Tab. \ref{main_results}.

\begin{table}[!ht]
\vspace{-0.5em}
    \caption{Node clustering results. $\triangle _x =0.01x$ denotes the Std. }
    \vspace{-0.8em}
    \centering
    \label{clustering}
    \begin{tabular}{l|cc|cc}
    \hline
        Model & \multicolumn{2}{|c|}{Photo} &\multicolumn{2}{|c}{Computers}  \\ \hline
        Metric & NMI & ARI & NMI & ARI \\ \hline
        GAE & 0.616±$\triangle_1$ & 0.494±$\triangle_1$ & 0.441±$\triangle_0$ & 0.258±$\triangle_0$ \\
        VGAE & 0.530±$\triangle_4$ & 0.373±$\triangle_4$ & 0.423±$\triangle_0$ & 0.238±$\triangle_0$ \\ 
        DGI & 0.376±$\triangle_3$ & 0.264±$\triangle_3$ & 0.318±$\triangle_2$ & 0.165±$\triangle_2$ \\ 
        MVGRL & 0.344±$\triangle_4$ & 0.239±$\triangle_4$ & 0.244±$\triangle_0$ & 0.141±$\triangle_0$ \\ 
        BGRL & 0.668±$\triangle_3$ & 0.547±$\triangle_4$ & 0.484±$\triangle_0$ & 0.295±$\triangle_0$ \\ 
        GCA & 0.614±$\triangle_0$ & 0.494±$\triangle_0$ & 0.426±$\triangle_0$ & 0.246±$\triangle_0$ \\ 
        DMoN & 0.633±$\triangle_0$ & - & 0.493±$\triangle_0$ & - \\ 
        HomoGCL & 0.671±$\triangle_2$ & 0.587±$\triangle_2$ & 0.534±$\triangle_0$ & \textbf{0.396±$\triangle_0$} \\ \hline
        MUX-GCL & \textbf{0.712}±$\triangle_1$ & \textbf{0.609}±$\triangle_1$ & \textbf{0.552}±$\triangle_0$ & 0.388±$\triangle_1$ \\ \hline
    \end{tabular}
    \vspace{-1em}
\end{table}

\vspace{-1em}
\subsection{Node Clustering} 
\vspace{-0.0em}
We further credit the performance gain in node clustering on Photo and Computers datasets to our design principles (see Tab. \ref{clustering}): The PAE module adheres intra-class nodes and alienates inter-class ones by assigning affinity scores, while the MPC module compacts the clusters by filtering out inconsistent information. Clusters thus preserve more consistency and have better defined boundaries.

\begin{table}[ht]
\vspace{-0.0em}
    \caption{Variants of PAE models (in node classification task)}
    \vspace{-0.8em}
    \label{DW_VGAE}
    \centering
    \begin{tabular}{lccc}
    \hline
        PAE method & Cora & Pubmed & Photo \\ \hline
        Node2Vec & 85.33 ± 0.37 & \textbf{86.94 ± 0.24} & 93.73 ± 0.04 \\ 
        VGAE & \textbf{85.43 ± 0.21} & 86.63 ± 0.15 & \textbf{93.89 ± 0.10} \\ \hline
    \end{tabular}
    \vspace{-0em}
\end{table}

\vspace{-0.5em}
\subsection{Ablation Study}
\vspace{-0.0em}We verify the effectiveness of the multiplex contrast mechanism and patch affinity estimation by testing the following variants: 
\begin{itemize}
    \item[(1)] \textbf{PAE}: only conducting same-scale contrast between the output embeddings, without engaging in cross-scale contrast.
    \item[(2)] \textbf{MPC}: performing a complete cross-scale contrast but refraining from utilizing patch affinity estimation to identify false negatives.
    \item[(3)] \textbf{PAE+MPC}: the full version of our model
\end{itemize}

As illustrated in Tab. \ref{tricks}, both PAE and MPC contribute to the performance gain, but with the optimal outcome attained when the two are integrated. This demonstrates that contrasting representations across scales and weighing false negatives both play crucial roles in preserving consistency information. We also remark that the results obtained by using either Node2Vec or VGAE for patch affinity estimation surpass those of existing SOTA models. 

\begin{table}[th]
\vspace{-0.5em}
    \caption{Ablation study (Acc (\%) ± Std for 5 seeds).}
    \vspace{-0.8em}
    \label{tricks}
    \centering
    \begin{tabular}{lccc}
    \hline
        Model$\backslash$Dataset & Cora & Citeseer & Photo \\ \hline
        w/o Both & 83.3 ± 0.4 & 72.1 ± 0.5 & 92.5 ± 0.2  \\
        PAE & 85.1 ± 0.3 & 73.3 ± 0.2 & 93.34 ± 0.09  \\
        MPC & 84.8 ± 0.4 & 73.4 ± 0.2 & 93.8 ± 0.1 \\ 
        PAE+MPC & \textbf{85.4 ± 0.2} & \textbf{73.8 ± 0.2} & \textbf{93.9 ± 0.1} \\ \hline
    \end{tabular}
\vspace{-1em}
\end{table}

\subsection{Runtime Analysis}\label{runtime}
\vspace{-0.2em}
We compare the training time of MUX-GCL with those of several advanced GCL methods (per epoch), as summarized in Tab. \ref{time}. Notably, compared to the computationally efficient GRACE, MUX-GCL improves considerably by increasing the training time only marginally. It is to remark that MUX-GCL is far more efficient than HomoGCL that computes saliency every epoch.

\vspace{-0.2em}
\begin{table}[h]
    \centering
    \caption{Time per epoch for GCL mehtods (on RTX 3090Ti)}
    \vspace{-0.8em}
    \label{time}
    \begin{tabular}{l|cccc}
    \hline
        Model & Cora & Citeseer & Photo & Computer \\ \hline
        GRACE & 0.20s & 0.02s & 0.05s & 0.12s \\
        ProGCL & 0.04s & 0.05s & 0.17s & 0.49s \\
        HomoGCL & 1.09s & 0.48s & 0.50s & 1.32s \\ 
        MA-GCL & 0.19s & 0.02s & 0.04s & 0.08s \\ \hline
        MUX-GCL & 0.04s & 0.05s & 0.16s & 0.42s \\ \hline
    \end{tabular}
\vspace{-1em}
\end{table}


\section{Conclusion}
We propose MUX-GCL, a novel cross-scale contrastive
learning paradigm, that grasps richer consistent information by utilizing multiplex representations as effective patches. Information contamination caused by conventional ways of constructing larger-scale subgraphs is mitigated in this framework. Commensurate to this paradigm, the scheme of patch affinity estimation is key to alleviate information loss from misjudging negative pairs of patches, which prevails in InfoNCE-based GCL methods. Notably, this affinity-informed mechanism is applicable to cross-scale contrasts, while all existing methods fail to be. Our approach is strictly proved theoretically and consolidated by its superior performance in downstream classification and clustering tasks relative to the SOTA GCL methods.

\bibliographystyle{IEEEtran}
\bibliography{IEEEabrv,reference}

\end{document}